\documentclass[pmlr]{jmlr}

\RequirePackage{graphicx}
 \usepackage{booktabs}
\usepackage{longtable}
\makeatletter
\def\set@curr@file#1{\def\@curr@file{#1}} 
\makeatother
\usepackage[load-configurations=version-1]{siunitx} 

\theorembodyfont{\upshape}
\theoremheaderfont{\scshape}
\theorempostheader{:}
\theoremsep{\newline}

\usepackage{graphicx}
\usepackage{xcolor}
\usepackage{amssymb}
\usepackage{multirow}
\usepackage{color, colortbl}
\usepackage{makecell}
\usepackage{gensymb}
\usepackage{mathtools}
\usepackage{algorithm,algpseudocode}
\usepackage{bm}
\usepackage{ragged2e}
\usepackage{booktabs}
\usepackage{amsmath,amsfonts}
\usepackage{textcomp}
\usepackage{cancel}
\usepackage{array}

\newcolumntype{C}[1]{>{\centering\arraybackslash}m{#1}}

\definecolor{TableGray}{gray}{0.9}
\definecolor{TableLightCyan}{rgb}{0.88,1,1}
\def\ours{\texttt{SimLVSeg}}

\usepackage{amssymb}
\usepackage{pifont}
\newcommand{\cmark}{\ding{51}}%
\newcommand{\xmark}{\ding{55}}%
\usepackage{subcaption}

\title[SimLVSeg]{SimLVSeg: Simplifying Left Ventricular Segmentation in Echocardiograms with Self- and Weakly-Supervised Learning}

\author{\Name{Fadillah Maani}
       \Email{fadillah.maani@mbzuai.ac.ae}\\ 
       \Name{Asim Ukaye}
       \Email{asim.ukaye@mbzuai.ac.ae}\\ 
       \Name{Nada Saadi}
       \Email{nada.saadi@mbzuai.ac.ae}\\ 
       \Name{Numan Saeed}
       \Email{numan.saeed@mbzuai.ac.ae}\\ 
       \Name{Mohammad Yaqub}
       \Email{mohammad.yaqub@mbzuai.ac.ae}\\ 
       \addr Department of Computer Vision\\
       Mohamed bin Zayed University of Artificial Intelligence\\
       Abu Dhabi, UAE} 

\begin{document}

\maketitle

\begin{abstract}
    Echocardiography has become an indispensable clinical imaging modality for general heart health assessment. From calculating biomarkers such as ejection fraction to the probability of a patient's heart failure, accurate segmentation of the heart structures allows doctors to assess the heart's condition and devise treatments with greater precision and accuracy. However, achieving accurate and reliable left ventricle segmentation is time-consuming and challenging due to different reasons. Hence, clinicians often rely on segmenting the left ventricular (LV) in two specific echocardiogram frames to make a diagnosis. This limited coverage in manual LV segmentation poses a challenge for developing automatic LV segmentation with high temporal consistency, as the resulting dataset is typically annotated sparsely. In response to this challenge, this work introduces \ours{}, a novel paradigm that enables video-based networks for consistent LV segmentation from sparsely annotated echocardiogram videos. \ours{} consists of \textit{self-supervised pre-training with temporal masking}, followed by \textit{weakly supervised learning} tailored for LV segmentation from sparse annotations. We demonstrate how \ours{} outperforms the state-of-the-art solutions by achieving a 93.32\% (95\%CI 93.21-93.43\%) dice score on the largest 2D+time echocardiography dataset (EchoNet-Dynamic) while being more efficient. \ours{}  is compatible with two types of video segmentation networks: 2D super image and 3D segmentation. To show the effectiveness of our approach, we provide extensive ablation studies, including pre-training settings and various deep learning backbones. We further conduct an out-of-distribution test to showcase \ours{}'s generalizability on unseen distribution (CAMUS dataset). The code is publicly available at \url{https://github.com/fadamsyah/SimLVSeg}.
\end{abstract}

\section{Introduction}
\label{sec:introduction}

Echocardiograms are a crucial modality in cardiovascular imaging due to their safety, availability, and high temporal resolution \cite{horgan2019echocardiography}. In clinical practice, echocardiogram information is used to diagnose heart conditions and understand the preoperative risks in patients with cardiovascular diseases \cite{ford2010systematic}. Through heartbeat sequences in echocardiogram videos, clinicians measure ejection fraction (EF) to assess the heart's capability to supply adequate oxygenated blood. The ejection fraction (EF) reflects the percentage of blood the heart can pump out of the left ventricular (LV), which is calculated as $(EDV-ESV)/EDV$ using the LV volume in the end-diastole (ED) phase end-systole (ES) phase. ED refers to the phase where the heart is maximally filled with blood just before contraction, while the ES phase happens immediately after the contraction where the volume of heart chambers is in its minimum stage. By accurately segmenting the heart structures, especially on the ED and ES frames, clinicians can assess the heart condition, detect any symptom, determine the appropriate treatment approach, and monitor the patient's response to therapy \cite{heidenreich2011forecasting}.

\begin{figure}[bt]
    \centering
    \includegraphics[width=0.8\textwidth]{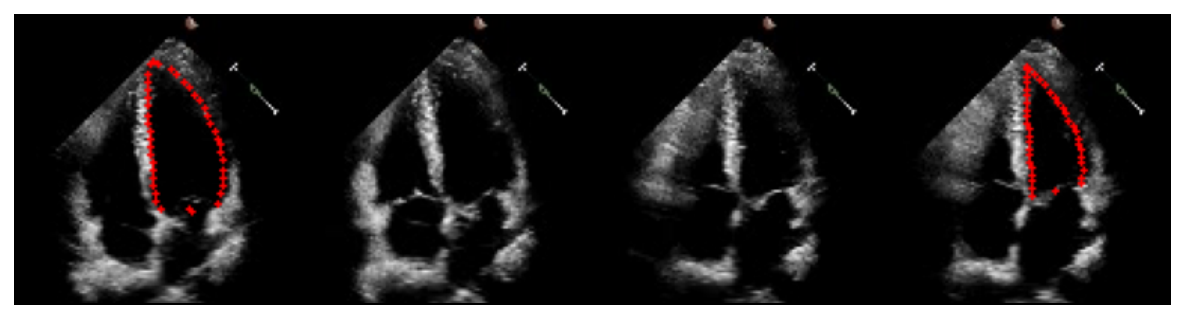}
    \caption{A sequence of an echocardiogram video \cite{ouyang2020video}. The number of frames varies, yet only two are labeled, i.e. the end-diastole (\textit{left-most}) and the end-systole (\textit{right-most}) frame. Annotators draw key points to represent the left ventricular (LV) region. Then, LV segmentation labels are inferred from the given key points.}
    \label{echo_dataset}
\end{figure}

The typical manual workflow of segmenting the LV is as follows: 1) a sonographer acquires an echocardiogram video using an ultrasound device and records the patient's heartbeat, 2) finds ED and ES by locating candidate frames indicated by the recorded heartbeat signal and then verifies them visually with the recorded echocardiogram video, 3) and ultimately draws some key points to represent the LV structure as shown in Figure \ref{echo_dataset}. That manual LV segmentation workflow is typically time-consuming and prone to intra- and inter-observer variability. The inherent speckle noise in echocardiograms makes the LV segmentation more challenging, as the LV boundaries are sometimes unclear. Hence, sonographers must consider the temporal context to eliminate the ambiguity caused by unclear heart structures in echocardiograms and perfectly segment the LV to achieve accurate results, which unfortunately means adding more burden for sonographers since they must go back and forth between echocardiogram frames to analyze the ambiguous boundaries properly. Automatic LV segmentation can help sonographers in solving this arduous task more efficiently.

A wide range of work on performing medical image segmentation using a supervised deep-learning approach is presented \cite{unet_2d, Isensee2021-yh}. The problem in echocardiogram segmentation, however, is more challenging since clinicians usually provide only two annotated frames per video, i.e. end-diastole (ED) and end-systole (ES) frames, resulting in limited labels for supervision. For instance, in EchoNet-Dynamic \cite{ouyang2020video}, the largest publicly available 2D+time echocardiography dataset, this utilizes less than 1.2 \% of the available frames when training in a 2D supervised setting. Consequently, early studies on the LV segmentation propose a frame-by-frame (2D) image segmentation solution \cite{2Dseg, bisenet, runet, ouyang2020video, weak_supervision}. These approaches do not capitalize on the periodicity and temporal consistency of the echocardiograms, which may lead to incoherence in the segmentation results from one frame to the next. In the worst-case scenario, the incoherence can lead to the ED and ES phase detection failure in the fully automatic ejection fraction prediction pipeline \cite{sarina_lightweight}. This has motivated a recent body of video-based echocardiogram segmentation approaches.

\cite{li2019recurrent} utilize a set of Conv-LSTM layers to ensure spatiotemporal consistency between consecutive frames. \cite{ahn2021multi} employ a multi-frame attention network to perform 3D segmentation. \cite{wu2022MiaTca} demonstrated the effectiveness of semi-supervision using mean-teacher networks and spatiotemporal fusion on segmentation. Recently, \cite{mclas} propose a two-stage training to enforce temporal consistency on a 3D U-Net by leveraging an echocardiogram ED \& ES sequence constraint. \cite{enforcedTemporal} improve the average segmentation performance by enforcing temporal smoothness as a post-processing step on video segmentation outputs.

These video-based approaches show high temporal consistency and state-of-the-art performance. However, they pose certain limitations. Recurrent units in \cite{li2019recurrent} incur a high computational cost. Multi-frame attention in \cite{ahn2021multi} similarly has computational cost correlated to the number of frames, and they are limited to using five frames. \cite{wu2022MiaTca} limit the temporal context to three frames to obtain optimum performance-compute trade-off. \cite{mclas} leverages a constraint in their training pipeline where the segmented area changes monotonically as the first input frame is ED and the last frame is ES in the same (\textit{one}) heartbeat cycle, thus limiting the usage of vastly unannotated frames in other cycles.

Investigating an alternative research direction, recent work has adopted a self-supervised learning (SSL) technique to effectively utilize the unannotated echocardiogram frames. \cite{saeed2022contrastive} use contrastive pre-training to provide self-supervision on echocardiograms. However, their solution uses frame-by-frame image segmentation with low temporal consistency. Recent studies in the natural domain, such as \cite{MaskedAutoencodersSpatiotemporal2022} and \cite{videomae}, adopt the masked autoencoders (MAE) for self-supervised pre-training to video networks, enabling accelerated training and show promising results in action recognition from natural videos.


The aforementioned works perform the LV segmentation from echocardiogram videos \textbf{either by} 1) analyzing frames independently with simple 2D deep learning models \textbf{or} 2) performing 2D+time analysis and developing models using complex training schemes. In our proposed method, while achieving state-of-the-art performance, we aim to mimic clinical assessment where doctors assess multiple frames concurrently in a simplified approach. Thus, we introduce \ours{} (\textit{\textbf{Sim}plified \textbf{LV} \textbf{Seg}mentation}), a novel training framework that enables video-based networks for LV segmentation, resulting in enhanced performance and higher temporal consistency. \ours{} consists of two training stages: self-supervised pre-training with temporal masking and weakly-supervised learning for LV segmentation, specifically designed to address the challenge of sparsely annotated (labeled) echocardiogram videos. Our main contributions are as follows:
\begin{itemize}
    \item We introduce a novel paradigm of performing LV segmentation with \ours{}. We show that it is feasible to develop video-based segmentation networks for LV despite the nature of sparsely annotated echocardiogram data. These networks effectively leverage spatial and temporal analysis, ensuring consistency across video frames. \ours{} is simple yet effective, opening new research directions for efficient and reliable LV segmentation empowered by video-based segmentation networks.
    \item We demonstrate how \ours{} outperforms the state-of-the-art in the LV segmentation on Echonet-Dynamic, the largest 2D+time echocardiography dataset, in terms of performance and efficiency through extensive ablation studies.
    \item We show \ours{}'s compatibility with two types of video segmentation networks: 2D super image \cite{fan2022can,sobirov2022segmentation} and 3D segmentation networks with various encoder backbones. This indicates that the excellent performance can be attributed to the \ours{} design rather than the selection of underlying network architectures.
\end{itemize}

\section{Methodology}

Our proposed method (\ours{}) is demonstrated in Figure \ref{fig_main}. \ours{} is composed of a \textit{self-supervised temporal masking} approach that leverages vastly unannotated echocardiogram frames to provide a better network initialization for the downstream LV segmentation task by learning the periodic nature of echocardiograms, and a \textit{weakly supervised training} that allows a video-based segmentation network to learn the LV segmentation from sparsely annotated (labeled) echocardiogram videos without any heartbeat cycle constraint. The network utilizes unannotated frames for a pre-training stage and learns from annotated frames in a weakly-supervised manner. The performance of the proposed method was evaluated with 3D segmentation and 2D super image (SI) segmentation \cite{fan2022can,sobirov2022segmentation} approach, as depicted in Figure \ref{3d_vs_2d-si}. The details are described below.

\subsection{Self-Supervised Temporal Masking.}

\begin{figure}[t]
    \centering
    \includegraphics[width=0.9\columnwidth]{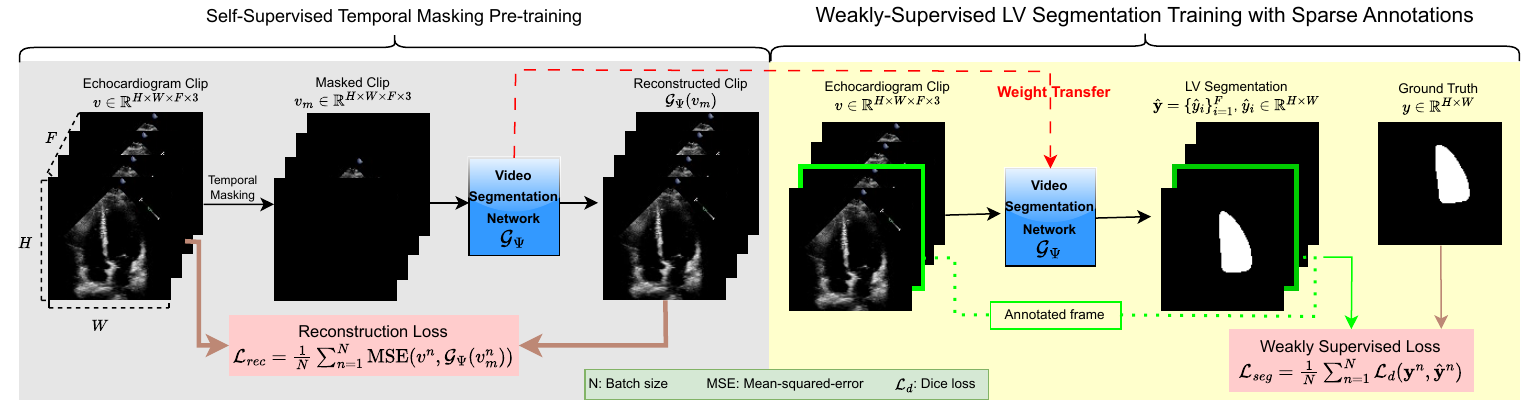}
    \caption{An illustration of \ours{}. A video segmentation network is developed to segment LV on every input echocardiogram frame. The network is pre-trained using a self-supervised temporal masking method, which is then fine-tuned on the LV segmentation task with sparse annotations.
    }
    \label{fig_main}
\end{figure}

In the EchoNet-Dynamic \cite{ouyang2020video} dataset, most of the frames are unannotated, thus the ability to perform supervised training is limited. To benefit from the vast amount of unlabeled frames, we implement a self-supervised temporal masking algorithm to pre-train our model. As depicted in Figure \ref{fig_main}, a clip of an echocardiogram video is retrieved, and a portion of the frames is masked. The model is then pre-trained to reconstruct the masked clip. Through this process, the model learns valuable latent information from the periodic nature of echocardiograms, e.g. the embedded temporal pattern or cardiac rhythm, that benefit the downstream LV segmentation task.

More formally, suppose $V$ is an echocardiogram video with $H \times W$ frame size. From $V$, we sample a clip $v \in \mathbb{R} ^{H\times W \times F \times 3}$ consisting of $F$ number of consecutive frames with a stride or sampling period of $T$. Then, we provide a masked clip $v_m \in \mathbb{R} ^{H\times W \times F \times 3}$ by randomly choosing $F_m$ number of frames ($F_m<F$) from $v$ and adjusting their pixel values to 0. A video network $\mathcal{G}_\Psi$ with a set of parameters $\Psi$ is then pre-trained to reconstruct $v$ from $v_m$. The network $\mathcal{G}_\Psi$ is optimized by minimizing the following objective:
\begin{equation}
    \mathcal{L}_{rec}=\frac{1}{N}\sum_{n=1}^N \text{MSE}(v^n, \mathcal{G}_\Psi(v_m^n))
\end{equation}
where $N$ is the batch size.

\subsection{Weakly Supervised LV Segmentation with Sparse Annotation}

The sparsely-annotated echocardiogram videos make the LV segmentation challenging as training a video segmentation model on EchoNet-Dynamic is not trivial. To tackle the issue, inspired by \cite{cciccek20163dunet}, we propose a training strategy to develop a video segmentation network specifically for LV. As illustrated in Figure \ref{fig_main}, the network takes in $F$ number of frames and segments the LV on each frame. Then, the loss is calculated and backpropagated only based on the prediction of frames having a segmentation label.

More formally, let $\mathcal{G}_\Psi$ be the pre-trained video segmentation network which takes in an input echocardiogram clip  $v \in \mathbb{R} ^{H\times W \times F \times C}$ and predicts LV segmentation $\hat{\bm{y}}=\{\hat{y}_1, \hat{y}_2, \ldots, \hat{y}_F\}$, $ \hat{y_i}\in \mathbb{R}^{H\times W}$, where $F$, $C$, and $H \times W$ are the number of frames, the number of channels which is 3, and the frame size, respectively. Also, let $\bm{y}=\{y_1, y_2, \ldots, y_F\},y_i\in \mathbb{R}^{H\times W}$ denote the sparse segmentation label of the input clip where most $y_i$ are empty. Thus, we construct $\bm{y}$ for every sample by using the following rule:
\begin{equation}
    y_i = \begin{cases}
        y_i & \text{if }i\text{-th frame is }\textit{labeled} \\
        \o & \text{otherwise}
    \end{cases}
\end{equation}
Thus, the total dice loss $\mathcal{L}_d$ for every sample $n$ can be formulated as:
\begin{equation}
\begin{aligned}
\mathcal{L}_d(\bm{y}^n, \bm{\hat{y}}^n) & = \sum_{i=1}^F \ell_d \left(y_i^n, \hat{y}_i^n \right) \\
& = \underbrace{\sum_{j\in\mathcal{F}_l^n} \ell_d \left(y_j^n, \hat{y_j}^n\right)}_\text{labeled (annotated) frames} + \underbrace{\sum_{k\in \{1,\ldots,F\} \backslash \mathcal{F}_l^n} \ell_d \left(y_k^n, \hat{y}_k^n\right)}_\text{unlabeled frames}
\end{aligned}
\end{equation}
where $\ell_d$ is the \textit{frame-wise} dice loss,  and $\mathcal{F}_l^n$ is the set of indices of labeled frames for the $n$-th sample. The gradient of $\mathcal{L}_d$ w.r.t. a parameter $\psi \in \Psi$ is given by:
\begin{equation}
    \frac{\partial\mathcal{L}_d}{\partial\psi}(\bm{y}^n, \bm{\hat{y}}^n) = \sum_{j\in\mathcal{F}_l^n} \frac{\partial\ell_d}{\partial\psi}\left(y_j^n, \hat{y}_j^n\right) + \sum_{k\in \{1,\ldots,F\} \backslash \mathcal{F}_l^n} \cancelto{\color{red}0}{\frac{\partial\ell_d}{\partial\psi}\left(y_k^n, \hat{y}_k^n\right)}
\end{equation}
where $\frac{\partial\ell_d}{\partial\psi}\left(y_k^n, \hat{y}_k^n\right)$ can be simply set to zero because the $k$-th frame is unlabeled, preventing the unlabeled frames from contributing to the gradients. \textbf{Since (1)} $\hat{y}_j \in \mathcal{G}_\Psi\left(v\right)$, and \textbf{(2)} $\mathcal{G}_\Psi$ typically consists of shared-weights operators (e.g. convolution and attention), \textbf{then}
\begin{equation}
    \frac{\partial\ell_d}{\partial\psi}\left(y_j^n, \hat{y}_j^n\right) \in \mathbb{R} \implies \sum_{j\in\mathcal{F}_l^n} \frac{\partial\ell_d}{\partial\psi}\left(y_j^n, \hat{y}_j^n\right) \in \mathbb{R} \implies \frac{\partial\mathcal{L}_d}{\partial\psi}(\bm{y}^n, \bm{\hat{y}}^n) \in \mathbb{R}
\end{equation}
for all parameters $\psi$ in $\Psi$. Thus, although a clip $v$ is partially labeled and gradients do not come from unlabeled frames, \textit{this framework can facilitate training for all $\mathcal{G}_\Psi$ parameters}. Ultimately, the total segmentation loss is given by:
\begin{equation}
    \mathcal{L}_{seg}=\frac{1}{N}\sum_{n=1}^N\mathcal{L}_d(\bm{y}^n, \bm{\hat{y}}^n)
\end{equation}

During training, a clip is randomly extracted around an annotated frame from every video with the specified number of frames $F$ and sampling period $T$, resulting in more variations and acting as a regularizer. In other words, there is only a segmentation mask for one frame on every clip. To reduce randomness during the evaluation step, a clip is extracted from each video where an annotated frame is at the center of the clip.

\begin{figure}[t]
    \centering
    \includegraphics[width=\textwidth]{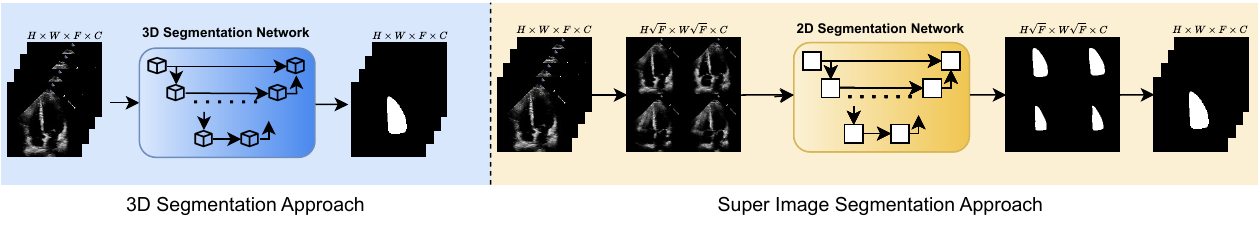}
    \caption{The 3D vs. 2D super image segmentation approach. The first approach utilizes a 3D segmentation network, while the second rearranges the echocardiogram clip as a super image and then utilizes a 2D network.}
    \label{3d_vs_2d-si}
\end{figure}

\subsection{Video Segmentation}
We aim to develop a video segmentation network $\mathcal{G}_\Psi$ capable of segmenting LV from an echocardiogram clip $v \in \mathbb{R} ^{H\times W \times F \times C}$. We consider two segmentation approaches as visualized in Fig. \ref{3d_vs_2d-si}, i.e. the 3D segmentation approach and the 2D super image (SI) approach. The 3D approach considers an echocardiogram clip as a 3D volume, while the SI approach addresses the video segmentation problem in a 2D fashion \cite{sobirov2022segmentation}. We describe the details of both approaches below.

\subsubsection{3D Segmentation Approach}
\begin{figure}[t]
    \centering
    \includegraphics[width=0.9\textwidth]{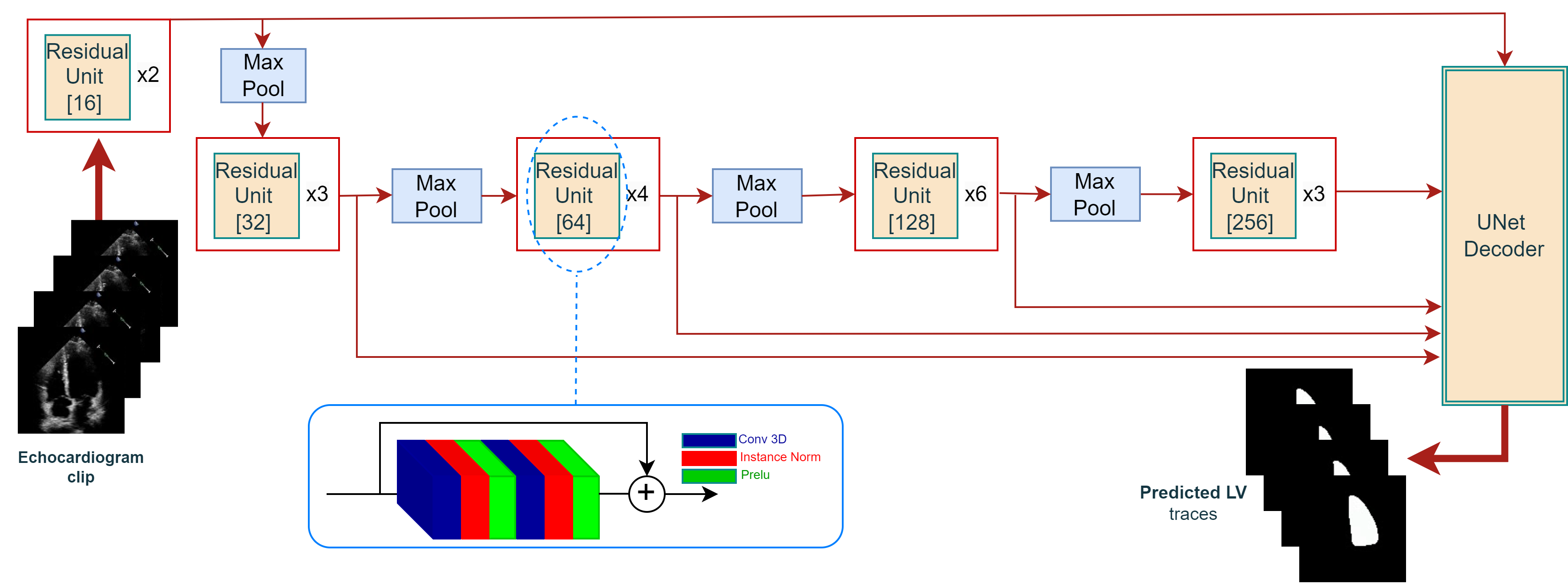}
    \caption{The 3D U-Net architecture. A residual unit \cite{residual_unet} consists of convolutional layers, instance norm layers, PReLU, and a skip connection. Residual Unit $[C]$ denotes a residual unit with $C$ number of feature channels.}
    \label{fig_3d_unet}
\end{figure}

Echocardiogram videos consist of stacked 2D images. Considering the time axis as the 3rd dimension allows 3D models to segment the LV on an echocardiogram clip. Thus, the 3D U-Net \cite{cciccek20163dunet} is utilized as the architecture. As depicted in Fig. \ref{fig_3d_unet}, we use a CNN with residual units \cite{residual_unet} as the encoder, which has 5 stages where the stage outputs are passed to the decoder. A residual unit comprises two Conv2D layers, two instance norm layers, two PReLU activation functions, and a skip connection.

\subsubsection{2D Super Image Approach}
An echocardiogram clip $v$ is rearranged into a single big image $x \in \mathbb{R} ^{\hat{H}\times \hat{W} \times C}$, where $\hat{H}$ and $\hat{W}$ are the height and width of the SI respectively. Since the SI works best with a grid layout \cite{fan2022can,sobirov2022segmentation}, we set the echocardiogram SI size to be $H\sqrt{F} \times W\sqrt{F}$. Hence, existing techniques for 2D image analysis can be well utilized to help solve the problem, e.g. state-of-the-art architectures, self-supervised methods, and strong pre-trained models.

The 2D U-Net \cite{unet_2d} is used as the main architecture with the UniFormer-S \cite{uniformer} as the encoder. We select the UniFormer-S since 1) it leverages the strong properties of convolution and attention, and 2) it is the recent state-of-the-art on EchoNet-Dynamic ejection fraction estimation \cite{echocotr}. In short, the network consists of 4 stages, where the first two stages utilize convolution operators to extract features, and the rest implement multi-head self-attention (MHSA) to learn global contexts. The inductive biases of convolution layers allow the model to learn efficiently and the MHSA has a large receptive field that is favorable for SI \cite{fan2022can}.

\section{Experimental Setup}
Experiments were mainly performed on EchoNet-Dynamic \cite{ouyang2020video}, a large-scale echocardiography dataset, using an NVIDIA RTX 6000 GPU with CUDA 11.7 and PyTorch 1.12. We additionally conducted an out-of-distribution (OOD) test of \ours{} on the CAMUS dataset \cite{camus}, a small echocardiography dataset, broadening the scope of our validation efforts.

\subsection{Dataset}

\subsubsection{EchoNet-Dynamic}

EchoNet-Dynamic \cite{ouyang2020video} is the largest publicly available 2D+Time echocardiograms of the apical four-chambers (A4C) view of the human heart. The dataset comprises approximately 10,030 heart echocardiogram videos with a fixed frame size of $112\times112$. Video length varies from 28 to 1002 frames, encompassing multiple heartbeat cycles, yet only two are annotated (ED \& ES frames). A sample echocardiogram sequence is given in Figure \ref{echo_dataset}.

To ensure a fair comparison with reported state-of-the-art methods, we adhered strictly to the organizer's provided split, consisting of 7460 training videos, 1288 validation videos, and 1276 test videos.

\subsubsection{CAMUS}

CAMUS \cite{camus} comprises 500 2D+time echocardiograms. Each echocardiogram captures a single heartbeat cycle with corresponding dense segmentation labels, i.e., segmentation annotations on all frames. The frame size varies, and the video length ranges from 10 to 42 frames, with a median of 20 frames. The ED and ES frames are located at the edges of the echocardiogram video, i.e. the first and last frames. This dataset also includes metadata associated with every echocardiogram scan, such as image quality, patient gender, and age.

\subsection{Implementation Details}

\subsubsection{Main Experiments}

We conducted our main experiments on the EchoNet-Dynamic dataset. We pre-trained our video segmentation models for 100 epochs with self-supervision. Each echocardiogram video was randomly sampled on every epoch with a specified number of frames ($F$) and a stride or sampling period ($T$) to give more variations. We utilized the AdamW optimizer and set the learning rate to 3e-4 learning rate and weight decay to 1e-5. A set of augmentations was applied to enrich the variation during training, consisting of color jitter, CLAHE, random rotation, padding to 124$\times$124 frame size, and random cropping to 112$\times$112. Then, the model is fine-tuned for the LV segmentation task with sparse annotations in a weakly-supervised manner for 70 epochs. Every video was sampled twice on every epoch to accommodate the annotated ED and ES frames. Hyper-parameters were set experimentally.

\subsubsection{Out-of-distribution Test}

We evaluated the OOD performance of \ours{} on the CAMUS dataset for LV segmentation. Each echocardiogram was resized to a 112$\times$112 frame to align with the EchoNet-Dynamic dataset frame size. For sequences with length shorter than $F$, we appended zero padding to the temporal axis from the end of the echocardiogram sequence. In contrast, for videos exceeding $F$ frames in length, we uniformly selected $F$ frames across the original sequence. This is achieved by calculating equally spaced indices over the sequence length and rounding these indices to ensure that they correspond to actual frame numbers. Each selected frame is then extracted to form a new sequence with exactly $F$ frames, ensuring the sequence length matches $F$. In addition, the OOD test was conducted using medium- and good-quality echocardiogram scans.

\section{Results}

\subsection{Comparison with the state-of-the-art}
\ours{} outperforms recent state-of-the-art approaches (\cite{ouyang2020video, Isensee2021-yh, weak_supervision}) on the EchoNet test set, as shown in Table \ref{tab_main} and Figure \ref{fig_comparisons}. We compare our method with the approach proposed by the EchoNet dataset publisher \cite{ouyang2020video}, the famous nnU-Net \cite{Isensee2021-yh}, which can perform better than specially designed echocardiography networks as mentioned in \cite{sarina_lightweight}, and the method achieving the highest dice similarity coefficient (DSC) on the test set \cite{weak_supervision}. \ours{} with 3D U-Net (\ours{}-3D) results in 93.32\% overall DSC, and \ours{}-SI approach shows on-par performance. Confidence interval (CI) analysis further shows no overlap between the 95\% CI of \ours{} with other state-of-the-art solutions, indicating that our improvements hold statistical significance over those methods with a \textit{p-value} of less than 0.05. The \ours{}-3D was trained with 32 frames sampled consecutively, while the \ours{}-SI was trained with 16 frames sampled at every 5\textsuperscript{th} frame. This experiment shows that a video segmentation network trained in a weakly-supervised manner is capable of segmenting the LV with a 3.8x lower computational cost compared to \cite{weak_supervision}.

\begin{table}[t!]
    \centering
    \begin{minipage}{0.65\textwidth}
    \caption{Dice similarity coefficient (DSC) on EchoNet-Dynamic test set. \ours{} shows state-of-the-art performance with fewer FLOPs and relatively fewer parameters. fvcore was utilized to count the FLOPs. Note that we report FLOPs on a per-frame basis (*).}\label{tab_main}
    {
    \resizebox{\textwidth}{!}{
    \begin{tabular}{l|l|l|l|r|r}
    \hline
    \rowcolor{TableGray} & \multicolumn{3}{|c|}{DSC (95\%CI)} & FLOPs & \#Params \\ \cline{2-4}
    \rowcolor{TableGray} \multirow{-2}{*}{Method}&  \multicolumn{1}{|c|}{Overall} & \multicolumn{1}{|c|}{ES} & \multicolumn{1}{|c|}{ED} & (G) & (M) \\
    \hline
    EchoNet-Dynamic & \multirow{2}{*}{92.00 \footnotesize (91.87-92.13)} & \multirow{2}{*}{90.68 \footnotesize (90.55-90.86)} & \multirow{2}{*}{92.78 \footnotesize (92.61-92.94)} & \multirow{2}{*}{7.84} & \multirow{2}{*}{39.64} \\
    \cite{ouyang2020video} & & & & &  \\
    \hline
    \makecell[l]{nnU-Net \\ \cite{Isensee2021-yh}} & 92.86 \footnotesize (92.74-92.98) & 91.63 \footnotesize (91.43-91.83) & 93.62 \footnotesize (93.48-93.76) & 2.30 & \textbf{7.37} \\
    \hline
    \makecell[l]{SepXception \\ \cite{weak_supervision}} & 92.90 - & 91.73 \footnotesize (91.54-91.92) & 93.64 \footnotesize (93.50-93.78) & 4.28 & 55.83  \\
    \hline
    \rowcolor{TableLightCyan}
    \ours{}-SI & 93.31 \footnotesize (93.19-93.43) & 92.26 \footnotesize (92.08-92.44) & \textbf{93.95} \footnotesize (93.81-94.09) & (*) 2.17 & 24.83 \\
    \rowcolor{TableLightCyan}
    \ours{}-3D & \textbf{93.32} \footnotesize (93.21-93.43) & \textbf{92.29} \footnotesize (92.11-92.47) & \textbf{93.95} \footnotesize (93.81-94.09) & (*) \textbf{1.13} & 18.83 \\
    \hline
    \end{tabular}
    }
    }
    \end{minipage}
    \begin{minipage}{0.3\textwidth}
    \centering
    \includegraphics[width=\columnwidth]{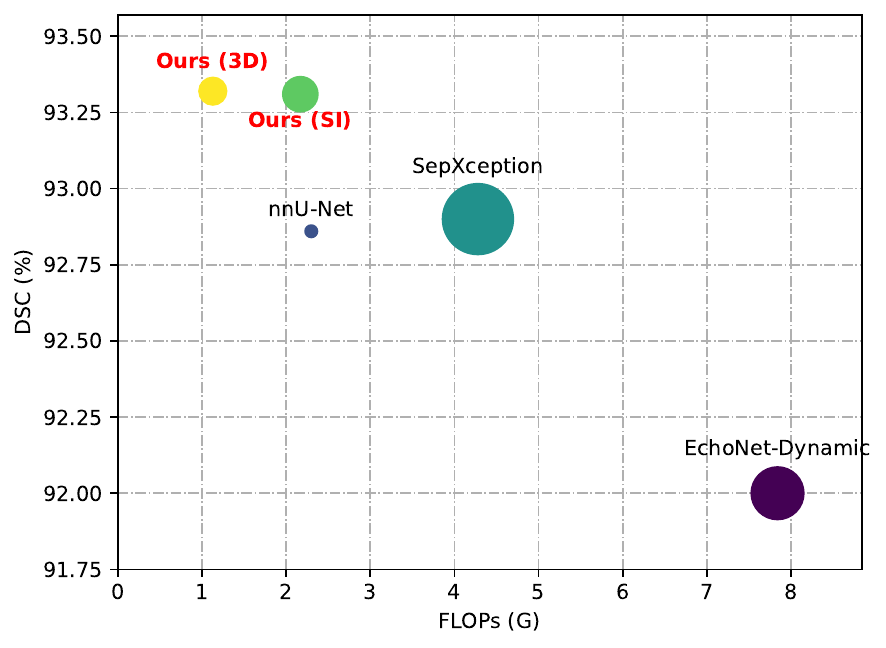}
    \captionof{figure}{A comparison with other SOTA solutions. The bubble size represents the number of parameters.}
    \label{fig_comparisons}
    \end{minipage}    
\end{table}

\begin{figure}[t]
    \centering
    \begin{minipage}[t!]{0.65\textwidth}
        \centering
        \begin{minipage}[t!]{0.49\textwidth}
            \centering
            \includegraphics[width=\textwidth]{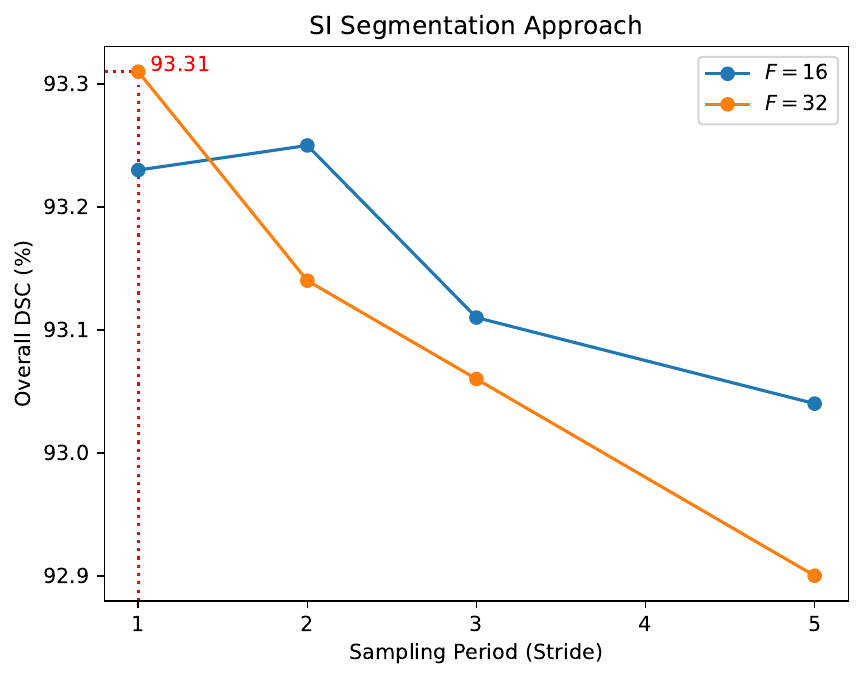}
        \end{minipage}
        \begin{minipage}[t!]{0.49\textwidth}
            \centering
            \includegraphics[width=\textwidth]{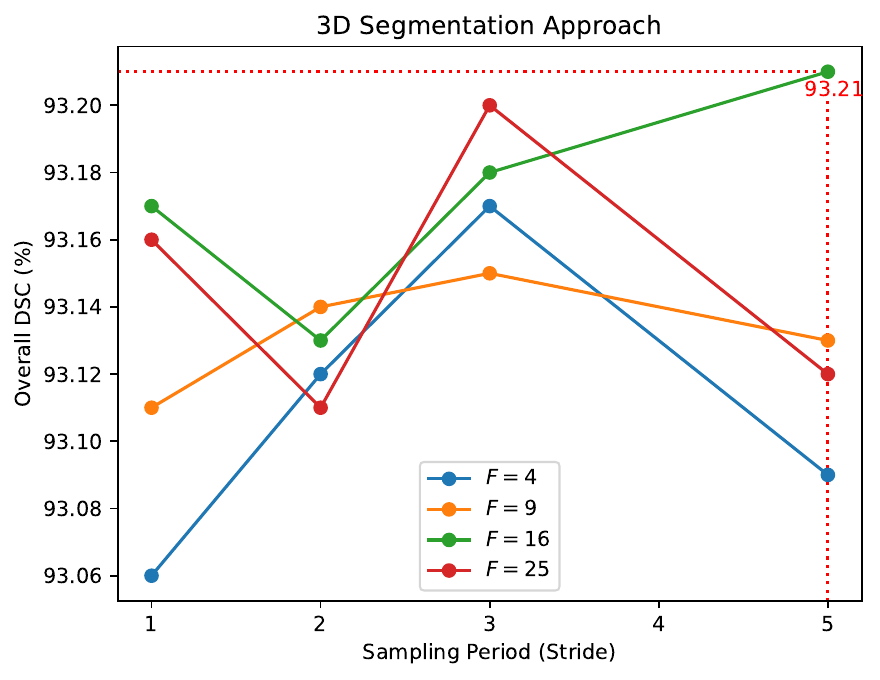}
        \end{minipage}
        \caption{Impact of the number of frames ($F$) and the sampling period ($T$). During this experiment, the UniFormer-S was pre-trained on ImageNet, and the 3D U-Net was trained from scratch.}
        \label{fig_n_frames_period}
    \end{minipage}
    \hspace{0.1em}
    \begin{minipage}[t!]{0.32\textwidth}
        \centering
        \includegraphics[width=\linewidth]{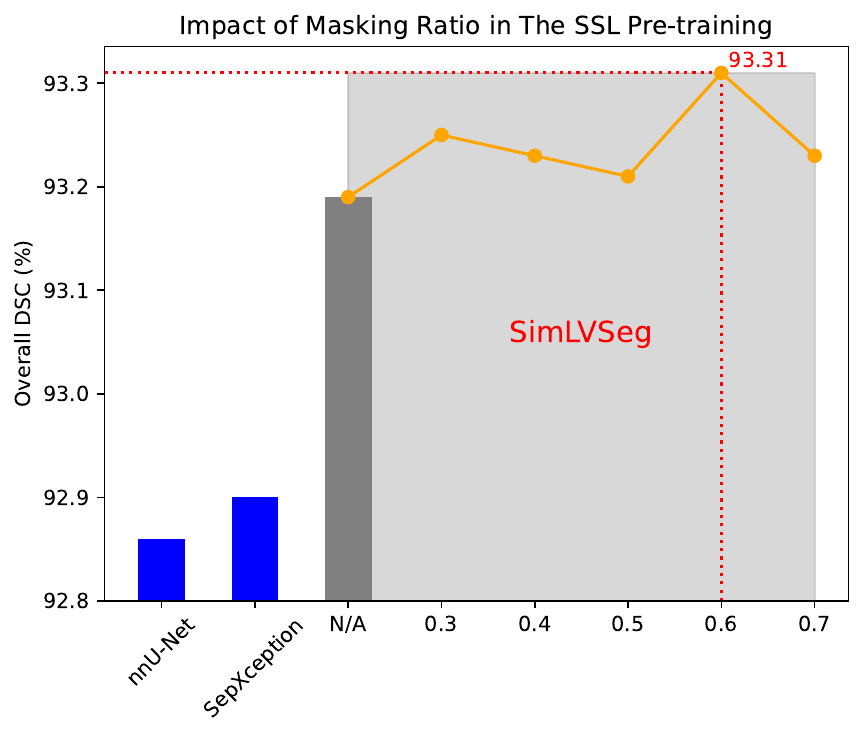}
        \captionof{figure}{Impact of masking ratio to \ours{}-SI. The optimum masking ratio is 60\%. N/A: w/o pre-training.}\label{fig_ssl}
    \end{minipage}
    
    \vspace{0.5em}
    
    \begin{minipage}{\textwidth}
        \centering
        \captionof{table}{An ablation study on various encoder backbones. Our approach is robust to the selection of backbone complexity. The SI backbones were pre-trained on the ImageNet dataset, while the 3D U-Net-S was trained from scratch.}\label{tab_backbones}
        \resizebox{0.9\textwidth}{!}{
        \begin{tabular}{l|r|r|r|r|r}
        \hline
        \rowcolor{TableGray} Approach & & \% DSC & & \multicolumn{2}{|c}{FLOPs (G)} \\ \cline{5-6}
        \rowcolor{TableGray}(\# Frames, Period) & \multirow{-2}{*}{Backbone} & (Overall) & \multirow{-2}{*}{Params (M)} & Single pass & One frame \\
        \hline
        Super Image (SI) & MobileNetV3 & 93.16 & \textbf{6.69} & \textbf{12.46} & \textbf{0.78} \\
        (16, 5) & ResNet-18 & 93.23 & 14.33 & 21.75 & 1.36 \\
        & ViT-B/16 & 92.98 & 89.10 & 120.20 & 7.51 \\
        \hline
        3D (32, 1) & 3D U-Net-S & \textbf{93.27} & 11.26 & 27.34 & 0.85 \\
        \hline
        \end{tabular}
        }
    \end{minipage}

    \centering
    \captionof{table}{\ours{}-3D performance on the CAMUS dataset (OOD). Using the proposed \textit{self-supervised temporal masking} for pre-training leads to better generalization.}\label{tab_ood}
    {
    \resizebox{0.9\textwidth}{!}{
    \begin{tabular}{c|l|l|l|l}
    \hline
    \rowcolor{TableGray} & \multicolumn{4}{|c}{DSC (95\%CI)}  \\
    \cline{2-5}
    \rowcolor{TableGray} \multirow{-2}{*}{SSL}&  \multicolumn{1}{|c|}{Overall} & \multicolumn{1}{|c|}{Middle} & \multicolumn{1}{|c|}{ES} & \multicolumn{1}{|c}{ED} \\
    \hline
    \xmark & 0.9044 \footnotesize (0.9039 - 0.9050) & 0.8976 \footnotesize (0.8952 - 0.8999) & 0.8901 \footnotesize (0.8875 - 0.8926) & \textbf{0.9234} \footnotesize (0.9217 - 0.9251) \\
    \rowcolor{TableLightCyan}
    \cmark & \textbf{0.9062} \footnotesize (0.9057 - 0.9067) & \textbf{0.9013} \footnotesize (0.8990 - 0.9034) & \textbf{0.8949} \footnotesize (0.8924 - 0.8973) & 0.9155 \footnotesize (0.9138 - 0.9172) \\
    \hline
    \end{tabular}
    }
    }
\end{figure}

\subsection{Ablation studies}

\subsubsection{Number of Frames and Sampling Period}
The number of frames $F$ and the sampling period $T$ play important roles \cite{echocotr, wu2022MiaTca}. Large $F$ allows a network to retrieve rich temporal information while increasing $T$ reduces redundancy between frames. We studied the combination of ($F$, $T$) to find the optimum pair as provided in Figure \ref{fig_n_frames_period}. The ($16$, $5$) combination results in the highest DSC of 93.21\% for SI while ($32$, $1$) gives the best performance for 3D approach, resulting in 93.31\% DSC. Additionally, all ($F$, $T$) pairs result in a better performance compared to the recent state-of-the-art \cite{weak_supervision}.

\subsubsection{SSL Temporal Masking}
\begin{figure}[H]
\centering
\subfigure[ ]{\label{fig:discussion_segmentation}\includegraphics[width=0.9\textwidth]{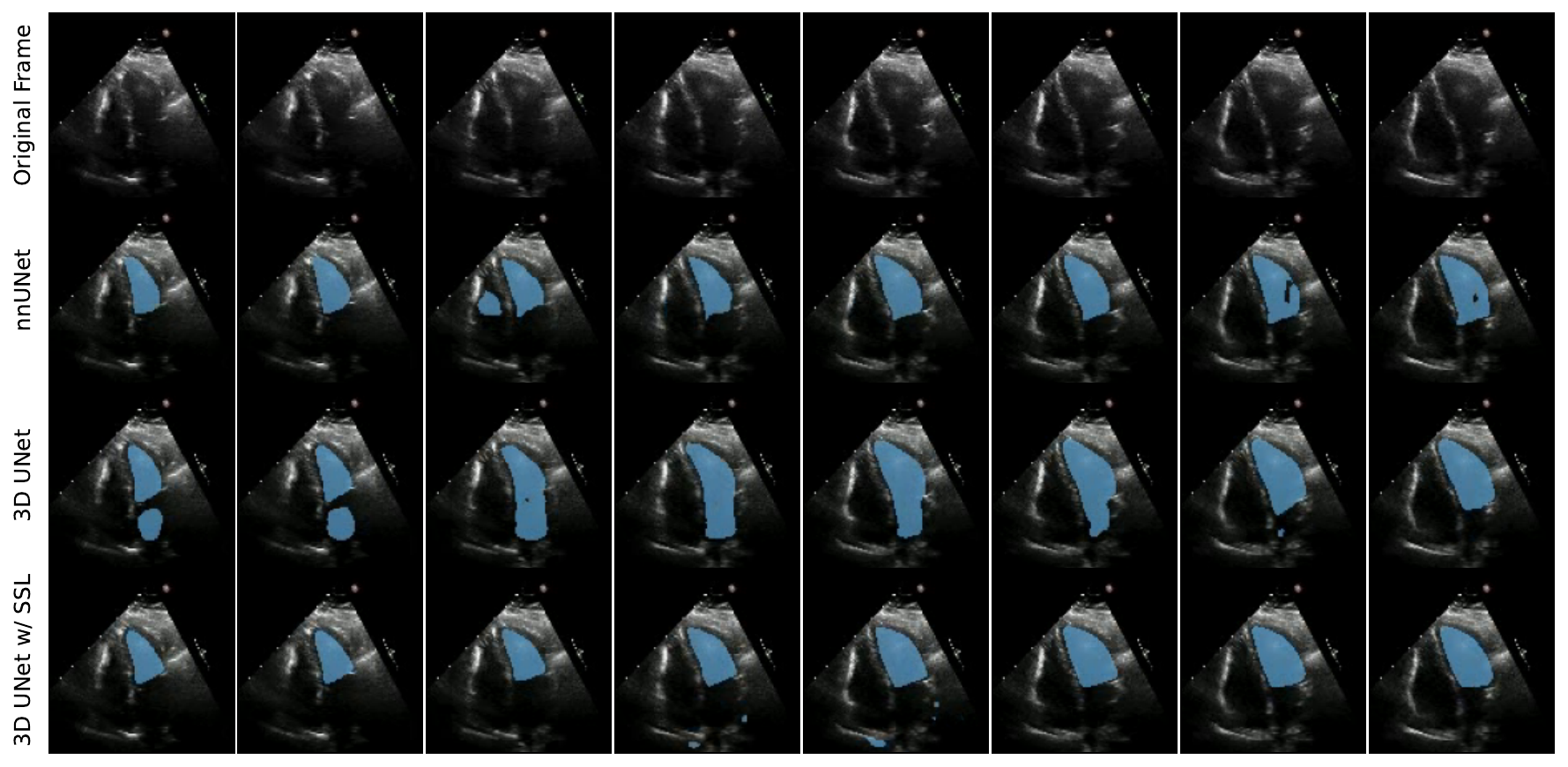}}
\subfigure[ ]{\label{fig:discussion_lv_area}\includegraphics[width=0.9\textwidth]{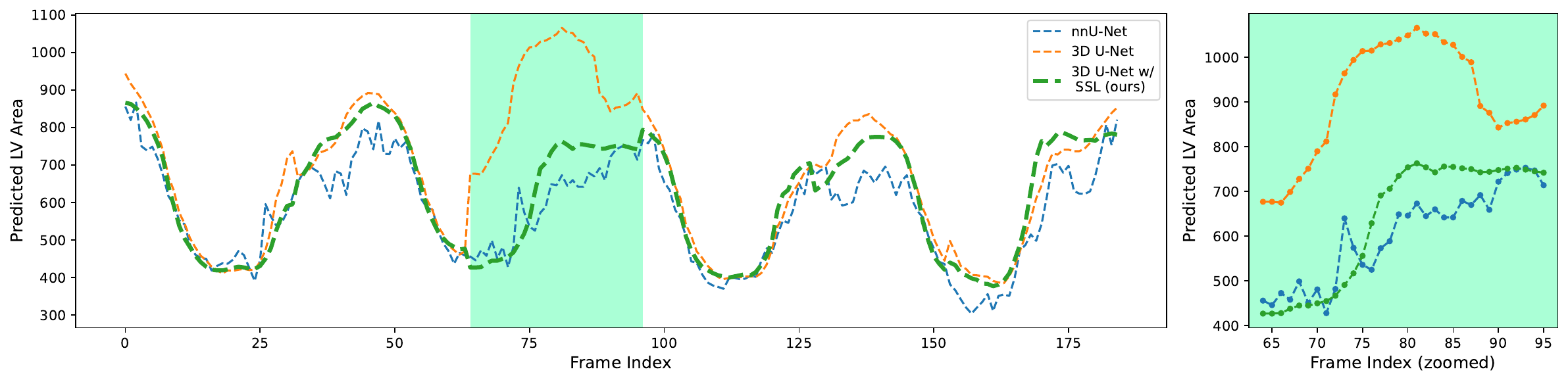}}
\subfigure[ ]{\label{fig:discussion_freq_comp}\includegraphics[width=0.8\textwidth]{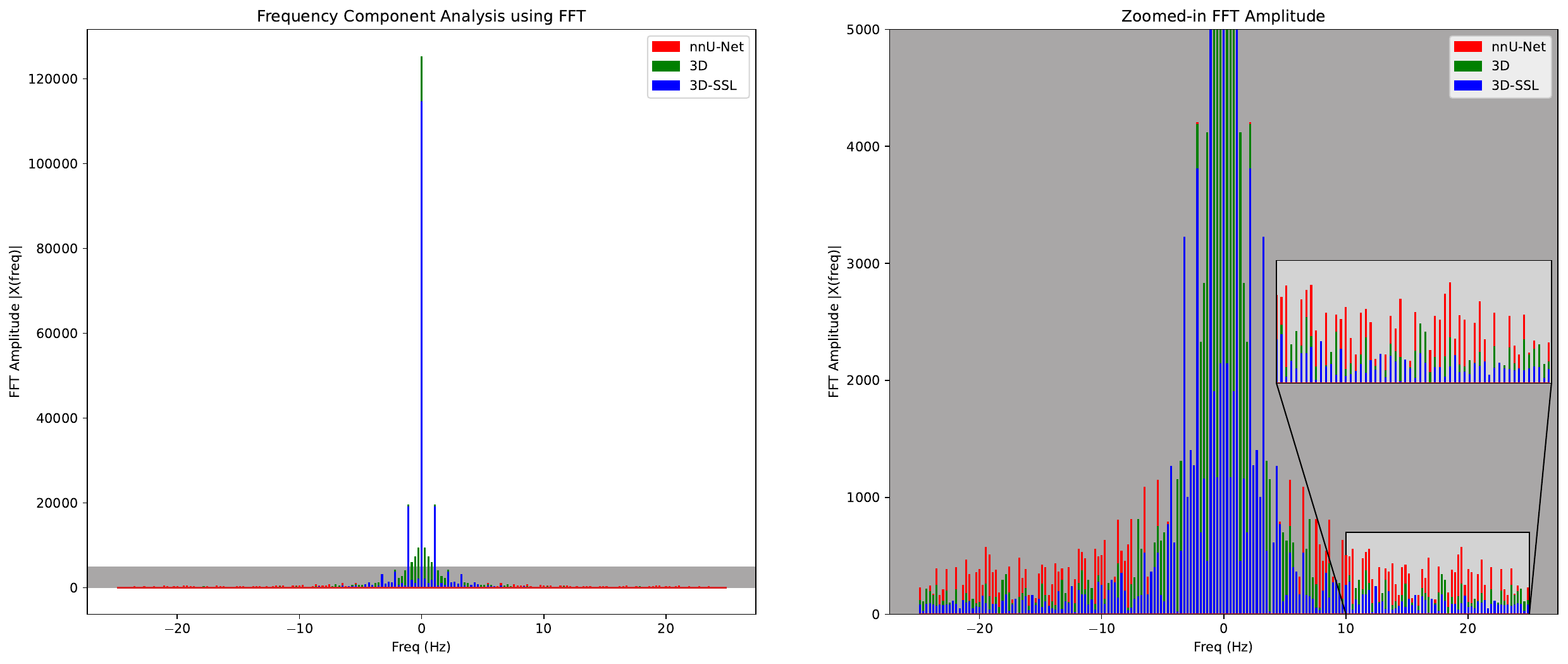}}
\caption{
Qualitative results for the performance of nnU-Net (2D) \cite{Isensee2021-yh}, 3D U-Net, and 3D U-Net with self-supervision (\ours{}-3D) without any post-processing trick against a challenging case where the Mitral valve is unclear. {(a)} We observe that the \ours{}-3D is more resilient to noise and missing artifacts in the input frames. {(b) The predicted LV segmentation area is smoother and more consistent for the \ours{}-3D compared to others. A zoomed version of the plot is shown on the right. (c) Frequency analysis using FFT on the predicted LV areas shows a lower magnitude of high-frequency components (i.e. lower noise) in the \ours{}-3D as compared to others.}
}
\label{fig:discussion}
\end{figure}

We conducted an ablation study (Figure \ref{fig_ssl}) to find the optimum value of the masking ratio and obtain the best results for 60 \% masking. We find that SSL pre-training helps maintain better temporal consistency and improve robustness (Fig. \ref{fig:discussion}). We verify the temporal consistency by analyzing the performance of the 3D U-Net when subjected to input video frames in random order and then comparing it to when the correct sequence is supplied.

\subsubsection{Different backbones}
An ablation study was performed on different encoders of the segmentation architecture to see how well our approach adapts to model complexity. We implemented ResNet-18 \cite{resnet}, MobileNet-V3 \cite{mobilenetv3}, and ViT-B/16 \cite{vit} as the encoder of the SI approach. We also tested with a smaller version of 3D U-Net (Fig. \ref{fig_3d_unet}), which consists of two residual units on every stage (3D U-Net-S). As provided in Table \ref{tab_backbones}, the experiment shows that the performance is robust to encoder backbones.

\subsubsection{OOD Test}

{

We conducted an additional test on the CAMUS dataset to determine if \ours{} can generalize well to samples with distributions unseen during training, such as different echocardiogram image contrasts, intensities, and original frame aspect ratios. Table \ref{tab_ood} presents the average DSC for all frames in the \textit{Overall} column, as CAMUS provides dense labels or annotations. Additionally, we present the DSC of the middle, ED, and ES frames separately. The \textit{self-supervised temporal masking} pre-training leads to improved overall performance with no overlap between the two 95\% CIs, indicating statistical significance. Moreover, the enhanced segmentation performance observed in the middle frame indicates that the SSL pre-trained model is particularly good at leveraging temporal dynamics, suggesting the model effectively utilizes the contextual information from the frames preceding and following the middle frame to improve its segmentation accuracy. Furthermore, while the ES DSC with SSL is higher, the ED DSC is lower than that of the network without SSL pre-training. In CAMUS, the ED and ES frames are located at the sequence edges, limiting temporal context from their preceding or following frames.
}

\section{Discussions}

Table \ref{tab_main} shows that while being more efficient, \ours{} outperforms the highest reported DSC on the EchoNet-Dynamic test set. \ours{} video networks aggregate both spatial and temporal information by analyzing multiple echocardiogram frames at a single pass. The networks predict an LV segmentation trace for every input frame at once, thus eliminating the redundancy in analyzing the same frames multiple times as in \cite{sarina_lightweight} and \cite{wu2022MiaTca}. In addition, \ours{} training pipeline is simple yet effective, easy to implement, and scalable, as it does not require pseudo labels \cite{clas, mclas} or temporal regularization \cite{enforcedTemporal}. Compared to \cite{clas, mclas}, \ours{} does not depend on a specific heart stage, thus eliminating the burden of locating the ED and ES frame when creating training data. This also allows us to easily leverage non-ED and -ES frames for supervision if their corresponding segmentation labels are available.
Figure \ref{fig_n_frames_period} highlights the robustness of \ours{} to the sampling hyperparameters. This allows for a broader design space to meet hardware limitations such as memory and compute power (FLOPs) while still achieving a satisfactory segmentation performance.

We observed that randomly masking a significant portion (60\%) of an echocardiogram clip during SSL pre-training results in the best performance. The masking SSL improves the overall DSC of the SI approach from 93.19\% to 93.31\%, as reported in Figure \ref{fig_ssl}. Further, as shown in Fig. \ref{fig:discussion}, we observe that self-supervision with temporal masking enables the network to maintain better temporal consistency across predictions in a given echocardiogram clip. Fig. \ref{fig:discussion_segmentation} demonstrates that a video segmentation model pre-trained with self-supervised temporal masking is more resilient to noise and missing artifacts. The pre-training stage also alleviates the over-segmentation and temporal inconsistency issues that are commonly encountered in echocardiography caused by unclear (or, even worse, invisible) boundaries. Also, the SSL pre-trained model achieves a smoother LV segmentation area prediction with significantly less rapid fluctuations (see Fig. \ref{fig:discussion_lv_area}), indicating better temporal consistency. We further investigate the phenomenon in the frequency domain by applying the Fast Fourier Transform (FFT) to the predicted signals (LV area) as depicted in Fig. \ref{fig:discussion_freq_comp}. We observe that SSL pre-training results in a lower magnitude of the high-frequency components, which are typically the result of noise and rapid fluctuation. Based on these observations, we hypothesize that \textit{the pre-training stage helps the 3D U-Net model to better learn the semantic features that are useful for estimating human heart structures in the A4C view}, resulting in a more robust prediction. Additionally, Table \ref{tab_ood} showcases the significant impact of the SSL pre-training stage when testing with samples subject to distribution shifts, indicating that the SSL pre-training enhances the model's generalization capability. These findings indicate that pre-training with self-supervision remarkably benefits the downstream LV segmentation task. Hence, self-supervised learning with vast echocardiogram videos can be a promising solution to provide strong pre-trained models that can generalize well in downstream echocardiography-related clinical tasks.

We have shown that both the SI and 3D segmentation networks trained using our proposed \ours{} are capable of accurately segmenting the left ventricle in echocardiogram videos. Both \ours{}-3D and \ours{}-SI outperform the state-of-the-art \cite{weak_supervision}, suggesting that the superior performance can be attributed to the \ours{} design rather than the selection of the underlying network architectures. The 3D U-Net performance is slightly better than the SI network with the UniFormer-S backbone. However, designing a backbone for 3D U-Net is not straightforward since it requires tedious hyperparameter tuning. On the other hand, there are plenty of optimized models that can be utilized as a backbone for the SI approach. For instance, MobileNetV3, with only 6.69 M of parameters, can give an on-par performance with 93.16\% overall DSC, as seen in Table \ref{tab_backbones}. The pre-trained models on ImageNet can also help generalize better if we only have a small amount of data. Moreover, many self-supervised learning algorithms for 2D can also be explored to further improve \ours{} performance.

\section{Conclusion and Future Work}
We propose a novel paradigm to tackle the LV segmentation task on echocardiogram videos, namely \ours{}. Our method outperforms other works on the EchoNet-Dynamic test set. \ours{} utilizes a video segmentation network that efficiently combines both spatial and temporal information. The network is pre-trained on a reconstruction task and then fine-tuned with sparse annotations to predict LV. An extensive experiment was performed to show the superiority of \ours{} both quantitatively and qualitatively. We expect that this work will motivate researchers to explore more about the video segmentation approach for LV instead of working on frame-by-frame prediction.

Despite \ours{}'s remarkable performance for consistent LV segmentation, we limited our experiments to self-supervision using temporal masking only. However, there remains scope to improve the self-supervision pre-training by identifying the optimum masking scheme between existing masking strategies \cite{videomae,wang2023videomaev2}, such as temporal, random spatiotemporal, space-wise, and block-wise masking. In addition, this work only considered LV segmentation from a single echocardiography view, i.e. A4C. Extending \ours {} for LV segmentation from multi-view echocardiogram videos can further improve the overall performance and its usage in clinical practice.

\newpage

\bibliography{sample}

\end{document}